\documentclass{article} 
\usepackage{iclr2025_conference,times}

\usepackage{amsmath,amsfonts,bm}

\def\eqref#1{equation~\ref{#1}}

\def\1{\bm{1}}

\DeclareMathAlphabet{\mathsfit}{\encodingdefault}{\sfdefault}{m}{sl}
\SetMathAlphabet{\mathsfit}{bold}{\encodingdefault}{\sfdefault}{bx}{n}

\usepackage{hyperref}
\usepackage{url}
\usepackage{graphicx}
\usepackage{multirow}
\usepackage{caption}  
\usepackage{fdsymbol}
\usepackage{subcaption}

\title{Ray-Tracing for Conditionally Activated Neural Networks}

\author{Claudio Gallicchio\\
Department of Computer Science\\
University of Pisa\\
Pisa, Italy \\
\texttt{\{claudio.gallicchio\}@unipi.it} \\
\And
Giuseppe Nuti\\
Silvretta Research, Inc. \\
Larchmont, NY \\
USA \\
\texttt{\{giuseppenuti\}@gmail.com} \\
}

\iclrfinalcopy 
\begin{document}

\maketitle

\begin{abstract}
In this paper, we introduce a novel architecture for conditionally activated neural networks combining a hierarchical construction of multiple Mixture of Experts (MoEs) layers with a sampling mechanism that progressively converges to an optimized configuration of expert activation. This methodology enables the dynamic unfolding of the network's architecture, facilitating efficient path-specific training.
Experimental results demonstrate that this approach achieves competitive accuracy compared to conventional baselines while significantly reducing the parameter count required for inference. Notably, this parameter reduction correlates with the complexity of the input patterns, a property naturally emerging from the network's operational dynamics without necessitating explicit auxiliary penalty functions.
\end{abstract}

\section{Introduction \& Background}

Mixture of Experts (MoEs) approaches have become the standard for large models, leveraging the principle of conditional activation to lessen the computational load; yet the conditionality is generally limited to a preset number of large blocks within a single layer of the network. The approach we propose implements a neural network where blocks (experts) are stacked over multiple layers. By expressing each block's output as the expected firing rate of a stochastic calculation path, we can simultaneously solve the inference and the selective activation problems. Importantly, since we model every block's output to be its expected activation rate, initiating a computational path from the input nodes or from within a block in the middle of the network will yield comparable results, allowing for a variety of new computational approaches, balancing the width- versus depth-first paradigm.\\

In broad terms, our aim is to create a network where blocks selectively activate depending on the input (i.e., different network regions will activate for different inputs) with the following properties:

\begin{enumerate}
\item The number of blocks (and neurons/synapses) computed and the total inference and training compute time increases with the difficulty of the input problem: harder inference problems require more computational resources and time;
\item The network presents a sequence of solutions, the first being the most approximate (but fastest), progressively becoming more precise (and more time-consuming).
\end{enumerate}

Partially activated networks have been studied utilizing numerous different solutions, loosely defined as the field of Conditional Computing: see 
\cite{han2021dynamic}
and \cite{doi:10.3233/IA-240035} for comprehensive surveys. With the advent of Large Language Models (LLMs), popularized by OpenAi's GPT, the number of network parameters has grown significantly. Models with billions -- and more recently trillions, e.g. \cite{fedus2022switch} -- are common, if not the standard for LLMs. The ability to compute and train such models on readily available hardware and/or further increases in the size of large networks hinges, in our opinion, on the ability to partially activate networks. Indeed, researchers at Google in \cite{fedus2022switch} leveraged a switch approach, the now ubiquitous MoEs, or Gated Networks, to significantly increase the number of parameters without a computational cost increase. Expanding this approach to a large number of experts, \cite{He2024MixtureOA} leverages a product-key approach to retrieve a sparse subset of experts. More recently, \cite{belcak2023fastfeedforwardnetworks} use a tree-like gate function at input level to segment the input space as a gate to multiple experts, ensuring activation of a single branch of each split by maximizing gate-level entropy. This work expands on \cite{raposo2024mixtureofdepthsdynamicallyallocatingcompute} -- generalizing the idea of skipping an expert to stacked experts.\\

\begin{figure}[tb]
    \centering
    \begin{subfigure}[t]{0.48\textwidth}
        \centering
        \includegraphics[width=1\linewidth]{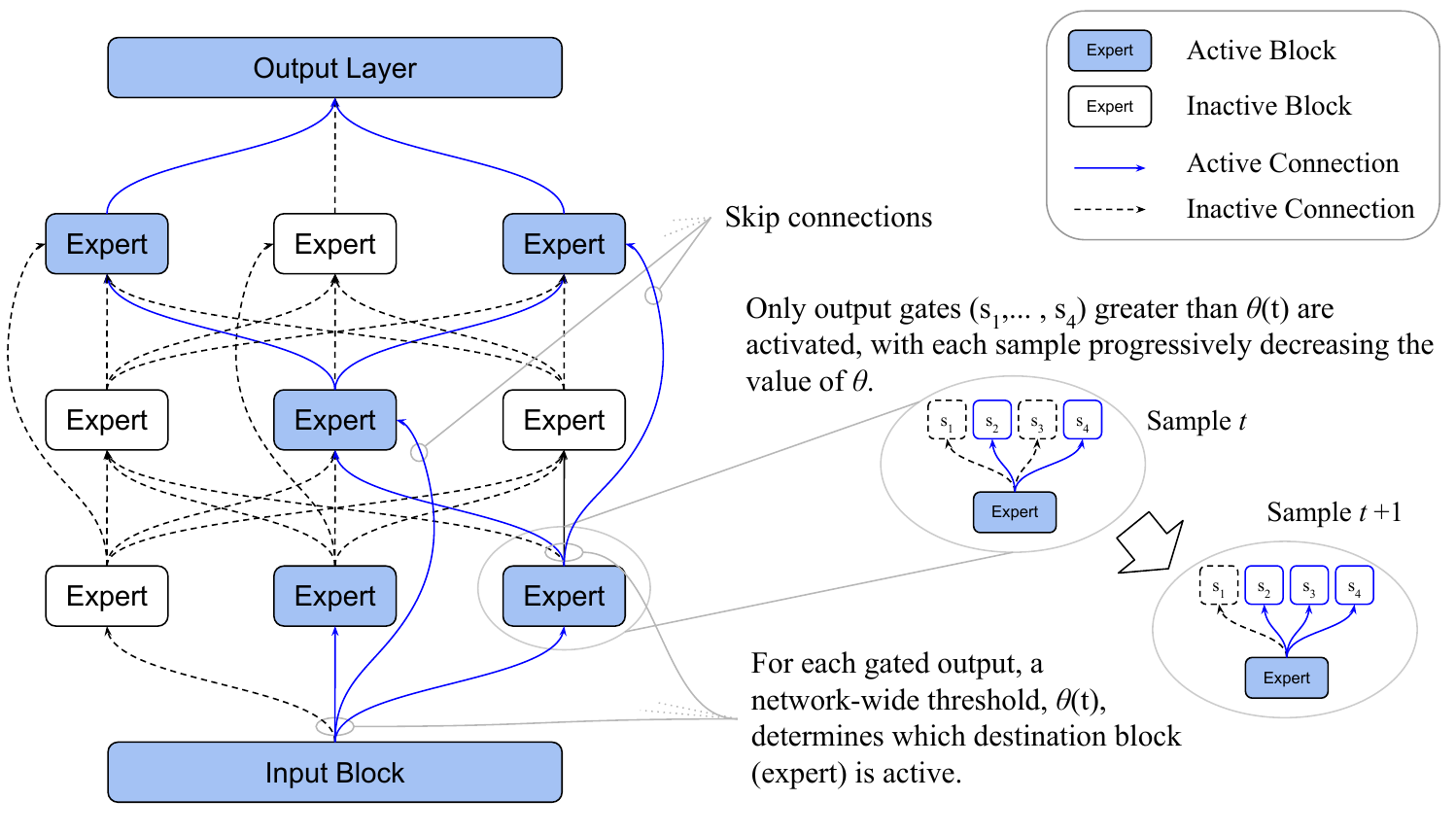}
        \caption{}
        \label{fig:sub1}
\end{subfigure}%
\hfill
\begin{subfigure}[t]{0.48\textwidth}
        \centering
        \includegraphics[width=1\linewidth]{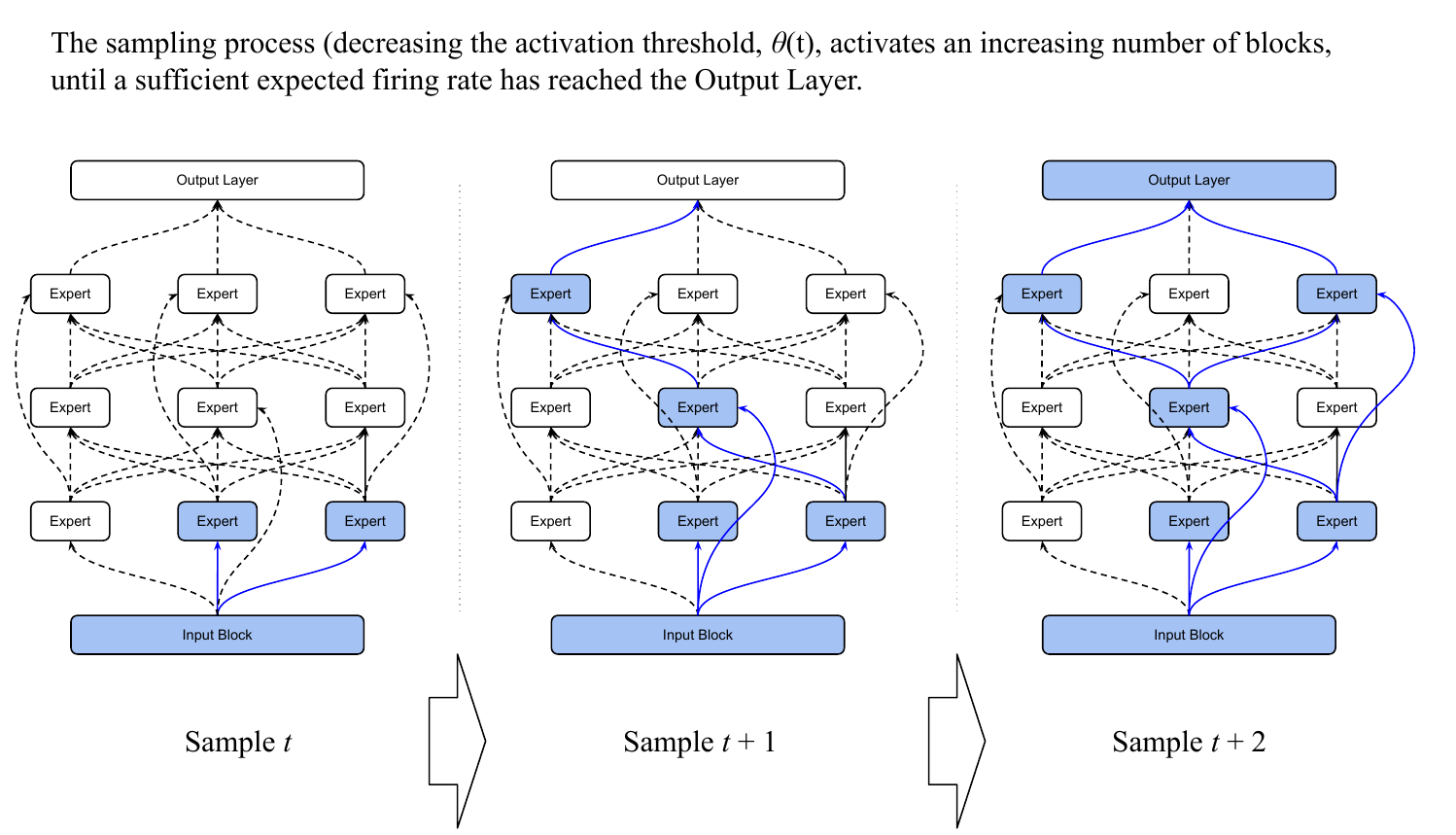}
        \caption{}
        \label{fig:sub2}
\end{subfigure}%
        \caption{(a) Each of the stacked blocks (experts) is activated when at least one of the incoming gates' value is above a threshold, $\theta$ (set at network level). (b) To sequentially sample approximations to inference problem, we start with a large value of $\theta$, progressively decreasing it (thus activating an increasing number of blocks) until a stopping condition has been met.}
        \label{fig.StackedMoEs}
\end{figure}

In this study, we explore networks for which the inference approach is done via multiple \emph{paths} which traverse the network activating blocks (or \emph{stacked experts}) independently -- in essence, balancing depth-first computations with the standard width-first approach. 
Our proposed model, which we refer to as \emph{RayTracing}, orchestrates the activation of these computational pathways, and is schematically illustrated in
Figure~\ref{fig.StackedMoEs}. In order to allow for information to build from the various inputs (and as a significant speed-up in the inference convergence), the output computed by a path is utilized (remembered) by any subsequent path traversing the same or neighboring block: in essence, the computation is an approximation which becomes progressively more precise as more paths activate the synapses, with the limiting case that an infinite number of paths would activate the entire network. The number of paths needed will depend on the stability of the approximation as more paths reach a specific target block -- thus providing a fast approximation that gets refined as more paths travel through the network. This property (a fast approximation followed by more precise, yet time consuming, solutions) can be valuable for time-sensitive decisions, such as self-driving cars and robotics in general.\\


Finally, in order to simultaneously train for the solution of the inference problem(s) and the computational efficiency, we enjoin the two problems into a single one: the output value and the probability of activation (or, equivalently, the activation rate), ensuring that the network is trained to output values that minimize the error w.r.t. the desired output and its activation. In other words, the objective function and the transformations from input into any of the output neurons include both the transformed output and the likelihood of the output neuron being activated. Importantly, multiple output neurons can be part of the same network (not necessarily mutually exclusive), selectively employing network resources based on inputs.


\section{Ray-Tracing Neural Networks}
\label{sec.model}

The output from each expert (or, neural block) is gated by using a threshold computation described as follows:
for each neural block $i$, we define its \emph{firing rate} as the sum of the incoming signals to that block, i.e., 
$r^{(i)} = \mathbf{z}^{(i) T} \mathbf{1}$, where $\mathbf{z}^{(i)}$ indicates the concatenation of the incoming signals into block $i$, 
and $\mathbf{1}$ is the vector of ones of the appropriate dimension.
In each block, computation takes place only when its firing rate is above a (network-level defined) threshold value, denoted as $\theta$. 
When the $i$-th neural block is activated it computes the output of a trainable neural module with $N$ output gates, denoted as $\mathcal{F}^{(i)}$. The gating mechanism can then be implemented by means of a $\textsc{Softmax}(\cdot)$ nonlinearity applied to the $N$ outgoing connections from the block. Formulaically, the output of the $i$-th block is computed as $\mathbf{s}^{(i)} = \textsc{ReLU}(r^{(i)} - \theta) \; \textsc{Softmax}(\mathcal{F}^{(i)}(\mathbf{z}^{(i)}))$, where the output of the ReLU nonlinearity has the effect of modulating the outgoing signals from the block, inhibiting the cases in which the block remains inactive.
Notice that potentially $\mathcal{F}^{(i)}(\cdot)$ can be cast to any neural operator, e.g., a multi-layer perceptron, a convolutional neural network, a residual block, etc.

Given this context, we introduce the two major characteristics of our proposal, namely (1) a \emph{hierarchical construction} of the resulting MoE architecture, and (2) a  process that unfolds the network \emph{through sampling}, thereby enabling effective training of the relevant paths within the architecture.
For (1), the neural blocks composing our MoE architecture are organized in layers, where each gated output from block is routed towards another block in a successive layer, or to the output layer. Notice that each outgoing output from a block can serve as input only for one other block in the architecture. The general architectural organization is illustrated in Figure~\ref{fig.StackedMoEs}.
Regarding (2), we introduce a relaxation process on the threshold $\theta$, whose value is progressively reduced whenever the total sum of incoming connections to the output layer is below a certain value, indicated as $\theta_{out}$. Accordingly, firing rates $r^{(i)}$ and outputs $\mathbf{s}^{(i)}$ pertaining to each neural block evolve in time during this process, according to the following equations:
\begin{equation}
\label{eq.innerblock}
\begin{array}{l}
r^{(i)}(t) = \mathbf{z}^{(i)}(t-1)^{T} \mathbf{1}\\
\mathbf{s}^{(i)}(t) = \textsc{ReLU}(r^{(i)}(t) - \theta(t)) \; 
                      \textsc{Softmax}(\mathcal{F}^{(i)}(\mathbf{z}^{(i)}(t-1))),
\end{array}
\end{equation}
where $\theta(t)$ denotes the (decreasing) value of the threshold at time $t$ of the relaxation process. Notice that while $\theta(t)$ decreases, larger portions of the neural architecture will be progressively explored (in the limit, with $\theta(t) = 0$, all the neural blocks would be computed).
Further observe that the external input to the network is kept constant during the threshold relaxation, and at each time step $t$ the input blocks are fed by the same input information, e.g., the same image (although additional input sampling strategies can be possible).
This mechanism in facts unfolds the neural architecture through the sampling / relaxation process, similarly to the way in which a Recurrent Neural Network is unfolded over time in sequence learning -- and unlike standard MoE models,  backpropagation only occurs through the activated blocks. At inference time, the output of the model is computed based on the activations of the blocks in the architecture at the end of the sampling process. Interestingly, at training time, the trainable weights in the $\mathcal{F}^{(i)}$ modules are adjusted by backpropagating the gradients \emph{through sampling}, enabling learning the formation of the appropriate neural pathways based on the computational task at hand.

\section{Experiments}
\label{sec.experiments}
\textbf{Experimental Settings.}
We demonstrate the RayTracing approach illustrated in Section~\ref{sec.model} on an extreme version of such construct: a network in which the neural operation performed by each expert is that of a simple  linear layer of trainable neurons, i.e, for each $i$: $\mathcal{F}^{(i)}(\cdot) = \textsc{Linear}_{\mathbf{W}^{(i)}, \mathbf{b}^{(i)}} (\cdot)$, with $\mathbf{W}^{(i)}$ and $\mathbf{b}^{(i)}$ denoting the corresponding weight matrix and bias vector, respectively.
There are numerous approaches to computing the sequence of path-wise activations (see \cite{Nuti_2023} and \cite{Nuti_2023_2} for further details), though here we focus on a decreasing threshold approach. We instantiate the RayTracing neural network (Figure~\ref{fig.StackedMoEs}) as follows: the input block is organized into a number of $6$ modules, where each module $j = 1, \ldots, 6$ computes its activation according to: $\mathbf{s}_{in}^{(j)} = N_{in}^{(j)} \textsc{Softmax}(\mathbf{W}_{in}^{(j)} \mathbf{x} + \mathbf{b}_{in}^{(j)})$, where $\mathbf{W}_{in}^{(j)}$ and $\mathbf{b}_{in}^{(j)}$ respectively denote the weight matrix and bias vector of the $j$-th input module, $\mathbf{x}$ is the vector of external input features, and $N_{in}^{(j)}$ is the number of neurons in the $j$-th input module (we set $N_{in}^{(j)} = 5j$). 
The input weight matrix in each input module is employed as a sparse matrix, with $1\%$ of non-zero weights. The hierarchical MoE architecture is organized into $L = 4$ hidden layers with $16$ expert per layer, where each expert employs a simple linear layer with $16$ neurons. The output layer is implemented as a dense classification layer with $\textsc{Softmax}$ nonlinearity for applications to multi-class classification problems. 
Each neuron in each of the input modules propagates its activation to one of the experts in the hierarchy, using random connections pointing to an expert in the first hidden layer with probability $p = 0.5$, and pointing to any deeper hidden layer (skip connections) with probability $p_{skip} = (1-p)/(L-1)$. The outgoing connections from each expert are modeled in a similar way, where the probability of establishing a connection pointing to an expert in the next layer is equal to $p$, and that of establishing a skip connection to a deeper layer or the output layer is equal to the remaining $(1-p)$ divided by the total number of deeper layers in the architecture (including the output layer). Connections from the $L$-th hidden layer connects exclusively to the output layer.
We use a threshold $\theta_{out} = 0.5$ at the output layer, and scheduled a decreasing process on the network-level threshold $\theta(t)$, such that $\theta(0) = 1.0$, and $\theta(t) = 0.9\, \theta(t-1)$. The network is trained using Adam with learning rate $0.001$ for $500$ epochs, and early stopping (on a validation set) with patience $10$.
\\

\noindent
\textbf{Datasets.} Our evaluation is conducted across four established image classification benchmarks: MNIST, Fashion MNIST, USPS, and CIFAR-10. These datasets provide a diverse array of conditions for testing the robustness and efficacy of our proposed RayTracing model. MNIST and USPS, consisting predominantly of grayscale images of handwritten digits, present a fundamental challenge in image classification. Fashion MNIST, serving as a direct substitute for the original MNIST with its collection of fashion item images, offers a nuanced variation in the simplicity of grayscale imaging. CIFAR-10, markedly distinct, features small color images across ten different object categories, introducing a higher level of complexity with its inclusion of RGB channels and a broader variety of image content. For each dataset, we used the original training / test splits, with a further 80\%/20\% division (70\%/30\% for USPS) of the original training data into training set and validation set. In all cases, the input features were extracted by using a simple convolutional neural network with 5 convolutional layers (3 for USPS) with kernel size of $5$ and a dense output layer, after a minimal training using Adam for $3$ epochs ($30$ epochs for CIFAR-10). The vector of input features $\mathbf{x}$ corresponding to each image was extracted by flattening the activation of the last convolutional layer of the resulting network.\\

\begin{figure}[ht]
  \begin{minipage}{0.5\textwidth}  
    \centering
    \includegraphics[width=0.8\linewidth]{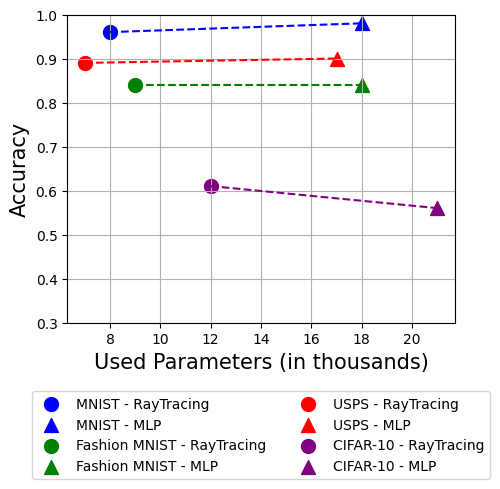}
  \end{minipage}\hfill
  \begin{minipage}{0.5\textwidth}  
    \caption{Scatter plot illustrating the trade-off between accuracy (evaluated on the test set) and the number of effectively used parameters (evaluated at inference time on the test set). 
    Different colors correspond to different datasets. Marker $\medblackcircle$ corresponds to Raytracing, $\medblacktriangleup$ corresponds to the baseline MLP. Points higher up in the plot correspond to better accuracy, points further to the left indicate greater computational efficiency. For RayTracing, we report the average number of used parameters across the entire test set.
    }
    \label{fig.comparison}
  \end{minipage}
\end{figure}

\noindent
\textbf{Results.}
In Figure~\ref{fig.comparison} we show the trade-off between the achieved level of accuracy and the number of used parameters on the test set across all the considered tasks.
Results on each dataset are averaged across $5$ repetitions, and are presented in comparison to the results obtained by a baseline multi-layer perceptron (MLP), trained in the same way as the RayTracing networks, and with a comparable number of total trainable parameters.
The results show that the proposed RayTracing methodology achieves classification performance comparable to or exceeding that of the baseline method, particularly in the case of CIFAR-10. Additionally, RayTracing significantly reduces the number of network parameters required for classification, with an average reduction of over 50\% compared to the baseline model.

\begin{figure}[ht]
  \centering
  \includegraphics[width=0.6\textwidth]{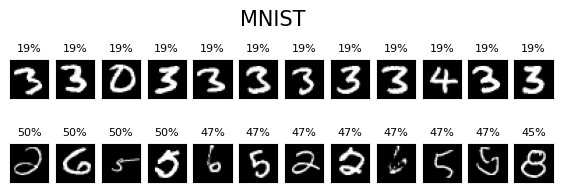}\\
  \includegraphics[width=0.6\textwidth]{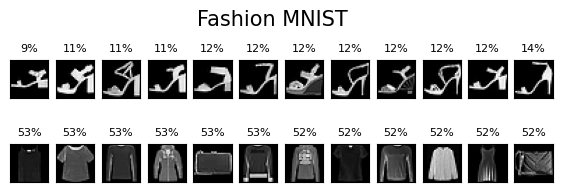}\\
  \includegraphics[width=0.6\textwidth]{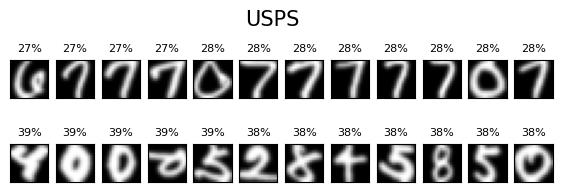}\\
  \includegraphics[width=0.6\textwidth]{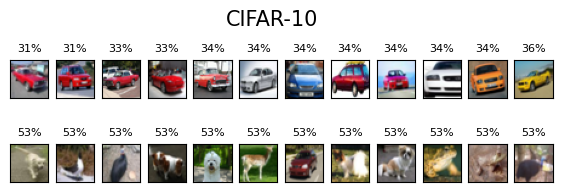}\\
  \caption{Input samples activating different proportions of the RayTracing neural network (indicated as a percentage value on top of each image).
  For each dataset, we present the 12 images that activated the least number of blocks (respective top rows) and the most number of blocks (bottom rows).}
  \label{fig.samples}
\end{figure}

To assess the ability of the RayTracing approach to selectively activate a different proportion of the network architecture, we report in Figure~\ref{fig.samples}  the percentage of active blocks for a selection of input images, resulting in the activation of a smaller (or larger) parts of the architecture used in the inference on the test set, for each dataset.
Easily recognizable images generally activate fewer blocks than images that are harder to recognize. 
For example, in MNIST, images representing well-written digits result in the activation of fewer blocks, whereas in Fashion MNIST, images corresponding to darker objects tend to activate more blocks. 
Similarly, classes that are easy to categorize tend to activate fewer blocks. 
For example, in CIFAR-10 automobiles represent easier classes to identify (see, e.g., \cite{MATLAB}), whilst birds, dogs, and cats tend to have the lowest accuracy. 
This matches our results in as much as the categories that require the most blocks are dogs and birds alongside some visibly unintelligible images, whilst the images that activate the least number of blocks are all automobiles.
Interestingly, we achieved these results 
without the use of any auxiliary penalty (such as for load-balancing, etc.). Though not explored in this study, we can favor faster calculations by adding a penalty term for node re-activations that are distant in sampling times, i.e. a term that inhibits paths which activate a neuron perviously activated (with the magnitude of the penalty proportional to the time between activations).

\section{Conclusions}

In this paper, we introduced a novel architecture for conditionally activated neural networks, referred to as RayTracing. The proposed method leverages a hierarchical Mixture of Experts (MoEs) structure, combined with a dynamic sampling mechanism, to facilitate efficient path-specific training. Our approach enables the selective activation of network blocks based on input complexity, optimizing both computational efficiency and classification performance.
Experimental results demonstrate that RayTracing achieves competitive accuracy compared to conventional multi-layer perceptron (MLP) baselines.
Moreover, our model significantly reduces the number of parameters required for inference, leading to a parameter reduction of over 50\% on average, without sacrificing accuracy. 

Future work may explore the application of RayTracing for time-sensitive tasks, such as robotics and autonomous systems, where efficiency is critical. Additionally, the extension of this approach to sequence learning, in combination with reservoir computing \cite{nakajima2021reservoir} and state-space models \cite{orvieto2023resurrecting, gu2023mamba}, could offer promising results. Another avenue for future research is the investigation of hardware-friendly implementations, which could further enhance the scalability and real-time applicability of the model.


\bibliography{RayTracing_iclr2025_workshop}
\bibliographystyle{iclr2025_conference}


\end{document}